\documentclass[letterpaper]{article} 
\usepackage{aaai2027}  
\usepackage[hyphens]{url}  
\usepackage{graphicx} 
\usepackage{natbib}  
\usepackage{caption} 
\usepackage{booktabs}   
\newcommand{\okoHus}{\d{o}k\d{o}}           
\newcommand{\okoVeh}{\d{o}k\d{\`o}}         
\newcommand{\okoHoe}{\d{o}k\d{\'o}}         
\newcommand{\yoruba}{Yor\`ub\'a}
\newcommand{\tonemetric}{DunDun}
\newcommand{\dundun}{d\`und\'un}             

\newcommand{\ANCHOR}{0.596}
\newcommand{\ANCHORCI}{95\% CI $[0.578, 0.614]$}

\title{Tone on a Budget: A Reference-Free Metric for Lexical Tone\\in Massively Multilingual Text-to-Speech}
\author{
    Moses Daudu\textsuperscript{\rm 1},
    Adeola Enitan Bamidele\textsuperscript{\rm 2},
    Honor-Jesus Bezaleel\textsuperscript{\rm 3}
}
\affiliations{
    \textsuperscript{\rm 1}Landmark University, Omu-Aran, Nigeria\\
    \textsuperscript{\rm 2}Federal University of Agriculture, Abeokuta, Nigeria\\
    \textsuperscript{\rm 3}Independent Researcher\\
    daudu.moses@lmu.edu.ng, bamideleae.23@student.funaab.edu.ng, bhonourjesus@gmail.com
}

\begin{document}

\maketitle

\begin{abstract}
In \yoruba{}, pitch alone separates \okoHus{} (husband, Mid), \okoVeh{} (vehicle, Low), and \okoHoe{} (hoe, High) --- the diacritics \emph{are} the tone marks. Yet character error rate (CER), the standard automated metric for text-to-speech (TTS), is in practice computed from ASR output that drops those marks: a synthesizer can ace CER and still say \emph{vehicle} for \emph{husband}. Automatic tone scoring is established in pronunciation training, where it grades a \emph{learner} against a known target and is trained or thresholded on labelled data; the protocols aimed at \emph{synthesized} speech instead score ASR output against tone labels, or put a phonetician in the loop. We carry that instrument over to TTS evaluation as \tonemetric{} --- named for the \dundun{}, the \yoruba{} talking drum that speaks through pitch alone --- an automated, \emph{reference-free} lexical-tone metric that needs no tone-labelled corpus and reports coverage beside accuracy, usable wherever the orthography marks tone: the gold High/Mid/Low sequence is read straight from the input text's diacritics (in TTS that text exists by construction, so no reference recording is needed), and the predicted sequence comes from the audio's pitch track (F0) via forced alignment. We validate the metric three ways. Flattening pitch with PSOLA resynthesis collapses \tonemetric{} while CER does not move. Inverting High and Low in the answer key of 300 native recordings drives the two-class readout to 0.14, symmetrically below its 0.35 chance level --- a consistency check on the scoring path rather than independent evidence, since that value follows algebraically from the correct-key result, as we show. And three native listeners, over 67 blind A/B trials, pick the tone-correct clip 89.6\% of the time (95\% CI 80.0--94.8; $p<10^{-4}$), confirming the manipulation is audible; whether \tonemetric{} tracks those judgements trial by trial is not resolved at this sample size. Applied to a massively multilingual zero-shot TTS model, \tonemetric{} shows what CER cannot: \yoruba{} tone sits near the native anchor before any \yoruba{} fine-tuning (0.567$\pm$0.02 over five decode seeds vs.\ 0.596; chance 0.33) --- despite the 21.4\% \yoruba{} CER the model's own paper reports --- and on our held-out set a few hours of clean audio halve the character error rate (same five decode seeds: 5.6\%$\to$2.7\% by 5\,h, 1.7\% by 15\,h) while tone --- already most of the way from chance to that anchor --- saturates within the hour. On non-tonal Swahili, CER already captures the gains: the metric a language needs is language-dependent. We release the metric and the complete validation protocol.
\end{abstract}



\section{Introduction}

\yoruba{} separates \okoHus{} (husband), \okoVeh{} (vehicle) and \okoHoe{} (hoe) by pitch alone --- the tone of the second syllable, Mid, Low or High --- and in \yoruba{} orthography the diacritics carrying that distinction \emph{are} the tone marks. A synthesizer putting the wrong pitch there has not produced an accent or an awkward prosody. It has said a different word.

The standard evaluation stack cannot see this. Character error rate (CER), the default automated intelligibility measure for text-to-speech (TTS), compares an ASR transcript of the synthesized audio against the input text --- and as computed in practice, that transcript carries no tone marks. A system flattening every tone therefore loses nothing measurable: \okoVeh{} transcribes as \okoHus{}, and CER scores a hit. The gap is not noise: to CER, a tone-flattened system and a tone-correct one are the same system.

This paper answers two questions. \textbf{Q1 (measurement):} what does a tone-specific axis show that CER cannot? \textbf{Q2 (data):} what does a few hours of clean, tone-marked audio actually buy?

\paragraph{Q1.} We introduce \tonemetric{} --- named for the \dundun{}, the \yoruba{} talking drum that speaks through pitch alone \citep{durojaye2021dundun} --- which scores a synthesized waveform against the tone-marked text it was asked to read. Gold High/Mid/Low comes from that text's diacritics, so no reference \emph{recording} is required; the prediction comes from the audio's own pitch track, windowed by forced alignment supplying timing and never a label. This grades free generations automatically, at checkpoint cadence.

\paragraph{Q2.} We fine-tune a massively multilingual zero-shot TTS model \citep{zhu2026omnivoice} on clean, tone-marked \yoruba{} at budgets of $0$, $1$, $5$, $15$ and $28.5$ hours, reading CER, \tonemetric{} and speaker similarity off one fixed $27$-utterance probe.

The answer is a dissociation. Reading one protocol end to end --- the five-decode-seed evaluation, the only one covering every budget --- intelligibility improves sharply and early: CER falls $5.6\% \to 3.4\% \to 2.7\% \to 1.7\%$ across $0$, $1$, $5$ and $15$\,h, halving by 5\,h and better than threefold by 15\,h. Tone over the same checkpoints starts near the top of its own axis and runs off it: $0.567$, $0.594$, $0.631$, $0.661$, against a chance level of $1/3$ and a native-speech anchor of \ANCHOR{} --- most of that range already covered before fine-tuning, and above the anchor from $5$\,h on. Speaker similarity stays flat ($0.81$--$0.84$); voice identity is not what changes. The zero-shot score already sits in the band the fine-tuned checkpoints occupy --- before any \yoruba{} fine-tuning, despite that model's own paper reporting $21.4\%$ \yoruba{} CER, its worst African language \citep{zhu2026omnivoice}. A CER-only evaluation would report a large, real win and say nothing about whether the tones were right.

Two kinds of error bar appear in that sweep and they answer different questions, so we never print them in one column or divide one by the other. \emph{Decode} variance resamples a single checkpoint at generation (five seeds); it is the only variance a $0$\,h condition can have, there being no training run to vary. \emph{Training} variance repeats the fine-tuning itself with the data held fixed, which is what a data-efficiency claim actually requires; we have it for the $1$, $5$ and $15$\,h budgets ($\pm0.025$, $\pm0.014$, $\pm0.056$ on tone) but not for $28.5$\,h, which is a single run. Read against training variance $5$\,h and $15$\,h overlap, and we do not claim $15$\,h beats $5$\,h.

One reading is a limitation of the metric, not of the model. Zero-shot tone, $0.567$, already sits within about one and a half decode-seed standard deviations of \ANCHOR{} (\ANCHORCI{}), the native-speech anchor measured through the identical pipeline; from $5$\,h onward the fine-tuned checkpoints score at or above it. That is not super-native tone. It means the axis saturates near \ANCHOR{} and stops ranking systems reliably there --- which is also why this sweep can show that the tone axis is nearly used up before fine-tuning begins while being unable to order its own largest budgets on tone.

\paragraph{Contributions.}
\begin{itemize}
\item \tonemetric{}, carrying the automatic tone verification built for pronunciation training over to TTS evaluation: a balanced tone accuracy reported with coverage, scoped to orthographies that mark tone reliably. Automated, reference-free tone scoring is not new, and on synthesized speech it exists via ASR or an expert labeller; to our knowledge \tonemetric{} is the first needing neither, and the first fit to a three-level African register system with no tone-labelled corpus.
\item Three validations, one a falsification oracle. Flattening pitch by resynthesis collapses \tonemetric{} while CER does not move. Inverting High and Low in the answer key of 300 native recordings --- predictions byte-identical, only the gold changed --- drives the two-class score to $0.14$, far below that readout's chance level of $0.35$, which no tone-blind score can imitate. And three native listeners over 67 blind A/B trials choose the tone-correct clip $89.6\%$ of the time. Whether \tonemetric{} tracks those judgements trial by trial is \emph{not} resolved at this sample size; we report that as a negative result.
\item A data-efficiency finding: intelligibility and tone dissociate under a fixed recipe, measured with training-seed rather than decode-seed error bars.
\item A non-tonal control: the same recipe on Swahili, where CER alone carries the result and \tonemetric{} is inapplicable by construction --- the orthography marks no tone.
\item A released toolkit: the metric, its frozen calibration, and the validation protocol.
\end{itemize}


\section{Related Work}

\paragraph{Tone-error-rate protocols.}
Tone error rate over tone-bearing units (TBUs) is established; three protocols sit closest. One scores synthesized Mizo speech, reference-free in our sense --- no reference \emph{recording}; in TTS the input text is given --- but Mizo orthography marks no tone, so a single expert phonetician supplies the gold \citep{mohanta2026mizo}. It reaches languages \tonemetric{} cannot, but not at checkpoint cadence. The other two are automated but ASR-mediated: a tone-aware error rate for \yoruba{} \emph{ASR} against a reference transcription \citep{chen2026linguistically}, and one on synthesized \emph{lip}-to-speech output, via ASR transcripts through a G2P \citep{lta2025l2s}. Both therefore score lexical recoverability, not acoustic realization --- a tone-blind front end still emits the right marked word from context. The latter carries a further cost: it presupposes a tone-emitting recognizer plus G2P, a stack existing for Mandarin and essentially no African tone language. \tonemetric{} keeps no recognizer \emph{transcript} in the loop: an acoustic model supplies alignment, never a label.

\paragraph{Reference-free evaluation of TTS prosody.}
Reference-free prosody scoring exists --- conditional prediction of discrete tokens \citep{ulgen2025ttscore}, prosodic ABX \citep{sun2026prosodic}, model-as-a-judge over expressiveness \citep{emergenttts2025} --- but each treats prosody as gradient and paralinguistic, emitting no categorical, word-identity-bearing label per TBU: none is built to register \okoVeh{} (vehicle) for \okoHus{} (husband) as an error. \tonemetric{} reports an (accuracy, coverage) pair over a three-way High/Mid/Low contrast with chance $1/3$; unvoiced or unalignable TBUs abstain rather than take free credit. It also does not reach $1.0$ on correctly-toned native speech: through the identical pipeline the native anchor is \ANCHOR{}, so the usable range is far narrower than $[1/3,1]$, and scores are read against that anchor, not as percentage-correct. What it admits, and no unconditional quality predictor can, is falsification by answer-key inversion.

\paragraph{Tone in learned and discrete representations.}
Lexical tone survives poorly in the discrete unit inventories modern TTS generates through, in \yoruba{} as in Mandarin, though it stays recoverable from continuous self-supervised (SSL) features \citep{osakuade2026lexical,bengono2024tone}. Those diagnose \emph{representations} of human speech; \tonemetric{} scores the synthesized waveform, downstream of the codec --- where alignment quality is the main failure mode, hence coverage beside accuracy. Our SSL probe is an anti-collapse check, not a tone cross-check: largely pitch-blind, and sharing an aligner with the F0 meter, so agreement is corroborative only.

\paragraph{African-language TTS and how it is scored.}
Tone-marked corpora \citep{meyer2022bibletts,gutkin2020yoruba,emezue2025naijavoices} and synthesizers speaking \yoruba{} zero-shot \citep{zhu2026omnivoice,qwen2026tts} exist; a tone axis does not. A recent 37-language low-resource benchmark blames African degradation on ``tonal contrasts and orthographic variation,'' yet scores with word error rate, a MOS predictor and native-listener MOS; on \yoruba{} the first tracks listeners ($|\rho| = 0.89$) while UTMOSv2 is flat ($|\rho| = 0.09$) \citep{guzman2026openbibletts}. Another reports ``tonal errors persisted,'' with nothing to isolate them \citep{edet2026efik}; and as synthetic speech now trains African ASR \citep{synthvoice2025}, a tone-blind metric seeds the next generation with tone errors. Our base model reports 21.4\% \yoruba{} CER, its worst African language \citep{zhu2026omnivoice}; that benchmark scores it at 10.33\% word error rate elsewhere \citep{guzman2026openbibletts}. Neither says whether the tones were right.

\paragraph{Automatic tone assessment in pronunciation training.}
The instrument is not ours. Computer-assisted pronunciation training (CAPT) has scored lexical tone automatically since at least 2006, by the recipe we use --- force-align against the known target text, take F0 per tone-bearing unit, normalize for register, threshold a tone or goodness-of-pronunciation score into correct/mispronounced \citep{zhang2006tone,elkheir2023review,tong2015got,li2018tone,wang2024pitchrnnt}. \emph{Automated}, \emph{reference-free} and \emph{lexical-tone} are that literature's, not ours, as is the minimal-pair argument we make with \okoHus{}/\okoVeh{}. What differs is not the pipeline. CAPT diagnoses a \emph{learner}, and to do so it fits a supervised tone model, a threshold derived from labelled data, or both \citep{tong2015got} --- data that is scarce for Mandarin \citep{wang2024pitchrnnt} and absent for \yoruba{}. \tonemetric{} needs no \emph{tone-labelled corpus}: its two decision thresholds were grid-searched once on 126 clips and frozen, and every gold label thereafter is read from orthography rather than annotated, so extending to a new tone-marking language costs no labelling at all. The difference is therefore one of degree in what is fitted and of kind in what must be annotated. CAPT rules every unit correct or mispronounced; \tonemetric{} abstains where F0 or alignment fails, reports coverage, and is read against a measured native anchor --- it scores a system, not a student. And it is calibrated for three level tones with downstep, not Mandarin contours. Pronunciation-error detection does reach synthesized speech \citep{wang2025voxevaluator}, but for segmental errors, supervised on an annotated corpus.

\paragraph{Components, and what we claim.}
We claim none of the parts --- forced alignment and the recognizers behind the CER baseline \citep{pratap2023mms,radford2022whisper}, PSOLA for the falsification oracle \citep{moulines1990psola}, a speaker-verification encoder \citep{desplanques2020ecapa}, an African-language SSL encoder for the anti-collapse probe \citep{alabi2024afrihubert} --- only their composition --- nor, per the paragraph above, the qualifiers \emph{automated}, \emph{reference-free} and \emph{lexical-tone}, which are CAPT's. Narrowly, then: automatic tone scoring of \emph{synthesized} speech exists, but ASR-mediated \citep{lta2025l2s} or expert-labelled \citep{mohanta2026mizo}; to our knowledge \tonemetric{} is the first needing neither, and the first fit to a three-level African register system without a tone-labelled corpus. The claim is scoped to orthographies marking tone \emph{reliably}: gold is the input diacritics, unmarked vowels count as Mid, so diacritized input is a precondition --- \yoruba{} here, not Hausa, only partially Igbo.%
\footnote{Unrelated to the Mande \emph{dunun} bass drum.}

\section{The \tonemetric{} Metric}

\tonemetric{} maps a waveform (24\,kHz here) and the \emph{intended} tone-marked text to a balanced tone accuracy and a coverage. No recording of the evaluated sentence is consulted, which is what lets it grade free generations.

\paragraph{Gold tones from orthography.}
Text is NFD-normalised, each base character grouped with its trailing marks; acute (U+0301), grave (U+0300) and macron (U+0304) give High, Low and Mid, an unmarked vowel gives Mid. Vowels are \{a,e,i,o,u\} plus \d{e},\d{o}, tested after stripping every U+0300--U+036F mark. A tone-bearing unit (TBU) is one orthographic unit: any vowel letter or tone-marked character.

\yoruba{} leaves Mid unmarked, so ``Mid'' and ``the transcriber omitted the diacritic'' are formally indistinguishable: under-marked text turns gold H and L into gold Mid, flattening the answer key in the very direction a flattened synthesiser fails. \tonemetric{} presupposes reliably diacritised input.

\paragraph{Locating TBUs.}
For alignment only the three tone marks are stripped (the sub-dot stays: \d{e}/\d{o}/\d{s} are aligner vocabulary entries), and each TBU's base-character index is recorded, aligning gold and predicted sequences by construction. Windows come from character-level CTC forced alignment against MMS log-probabilities (\texttt{mms-1b-all}, \texttt{yor} adapter, 16\,kHz) \citep{pratap2023mms}; out-of-vocabulary characters abstain individually, and alignment failure or a length mismatch falls back to equal-width windows over the voiced span, a path whose rate we report.

An acoustic model is therefore in the loop. It supplies time alignment only and never a label, so no recognizer \emph{transcript} enters the score, unlike ASR-mediated tone-error rates \citep{lta2025l2s,chen2026linguistically}. It is also the main failure mode: coverage says a window exists, not that it sits on the right syllable, and a misplaced window scores a real F0 value against the wrong gold label, invisibly. Alignment is validated on native read speech and only assumed to transfer to synthesized speech, so codec-independence rests on that assumption, not on coverage.

\paragraph{Pitch.}
F0 is tracked once per clip (SwiftF0, pyworld fallback; 65--400\,Hz, voicing confidence $0.9$, unvoiced frames NaN). A TBU's F0 is the median of its voiced frames over the late half of its window, $[t_0+0.5(t_1{-}t_0),\,t_1)$, where \yoruba{} tone targets are realised; an unvoiced late half retries the whole window before abstaining. Values become semitones (the reference cancels below).

\paragraph{Declination and register.}
\yoruba{} F0 declines across an utterance --- High falls several semitones from start to end under downdrift \citep{laniran2003downstep}, at a corpus-level rate of roughly $-12$\,Hz/s \citep{vanniekerk2012pitch}, so a late High can sit below an early Mid and a classifier banding raw F0 against a global median scores near chance on \emph{correct} speech. We therefore fit a Theil--Sen slope to (semitone, time) over all valued TBUs, using no tone labels, clamp it to $[-2,0]$\,st/s and subtract it (fewer than three valued TBUs keeps a zero slope). Residuals are re-centred on their per-utterance median: label-free, but not assumption-free, since it assumes Mid is the modal tone \emph{within the clip} --- an H- or L-heavy carrier slides it into that cluster, shifting every label there. Then $d_i\ge\theta_h\Rightarrow$\,H, $d_i\le-\theta_l\Rightarrow$\,L, else M.

\paragraph{Calibration.}
$\theta_h,\theta_l$ were fitted once and frozen. From OpenSLR-86 \citep{gutkin2020yoruba}, 380 clips of 6--16 TBUs were split sentence-disjointly (transcript MD5) into 126 calibration and 254 test clips. On the calibration split, a $6\times6$ grid over $\{0.5,0.75,1.0,1.25,1.5,2.0\}$\,st maximised balanced accuracy $-\,0.5\,|\mathrm{recall}_H{-}\mathrm{recall}_L|$, rejecting pairs whose label-permutation control exceeded $0.40$; $\theta_h{=}1.0$, $\theta_l{=}1.25$ won. That the winning pair is asymmetric, and of this magnitude, is consistent with reported category boundaries in African tone perception \citep{carterenyi2016contour}; we fit rather than assume it. Frozen, they read $0.580$ at coverage $0.917$ on the disjoint test split, against $0.334$ for that split's permutation control.

\paragraph{Aggregation, chance, and ceiling.}
A TBU abstains when its window is missing or unvoiced; abstentions are never scored ``M''. \tonemetric{} is the unweighted mean of the per-class recalls of tones with at least one covered TBU. Pooled over an evaluation pass all three tones are present and $\mathrm{recall}_c{=}P(\hat{y}{=}c)$ for any label-independent predictor, so pooled chance is exactly $1/3$ whatever the tone marginals; an all-Mid predictor measures $0.333$. Restricted to a subset $\mathcal{C}$ of $k$ tones --- a clip containing only those, or the two-class readout of the next section --- the mean runs over $\mathcal{C}$ alone and chance is $\sum_{c\in\mathcal{C}}P(\hat{y}{=}c)/k$. Restricting the gold does not restrict the prediction, so this reaches $1/k$ only for a predictor that never answers outside $\mathcal{C}$: a carrier holding only H and M has chance $(1-P(\hat{y}{=}\mathrm{L}))/2 \le 1/2$.

The statistic ranges over $[0,1]$. $1/3$ is not a floor but the expected value of a label-independent predictor: systematically \emph{inverted} predictions fall below it, which is what makes the answer-key oracle of the next section informative rather than merely a degradation test. The practical upper end is \ANCHOR{} (\ANCHORCI{}), a native-speech anchor measured through the same pipeline on 300 native recordings from 37 speakers that no trained model is evaluated on; per-class recall there is $0.56/0.67/0.56$ for H/M/L at coverage $0.92$. \tonemetric{} does not reach $1.0$ on correctly-toned native speech, so the usable range is far narrower than $[1/3,1]$; scores are read against \ANCHOR{}, and one at or above it means band saturation, not super-native tone. Coverage travels with every score, per class too, since abstention is not class-independent: final Low is often creaky. Pooled coverage below roughly $0.7$ signals failure whatever the accuracy.

\begin{figure}[t]
\centering
\includegraphics[width=\columnwidth]{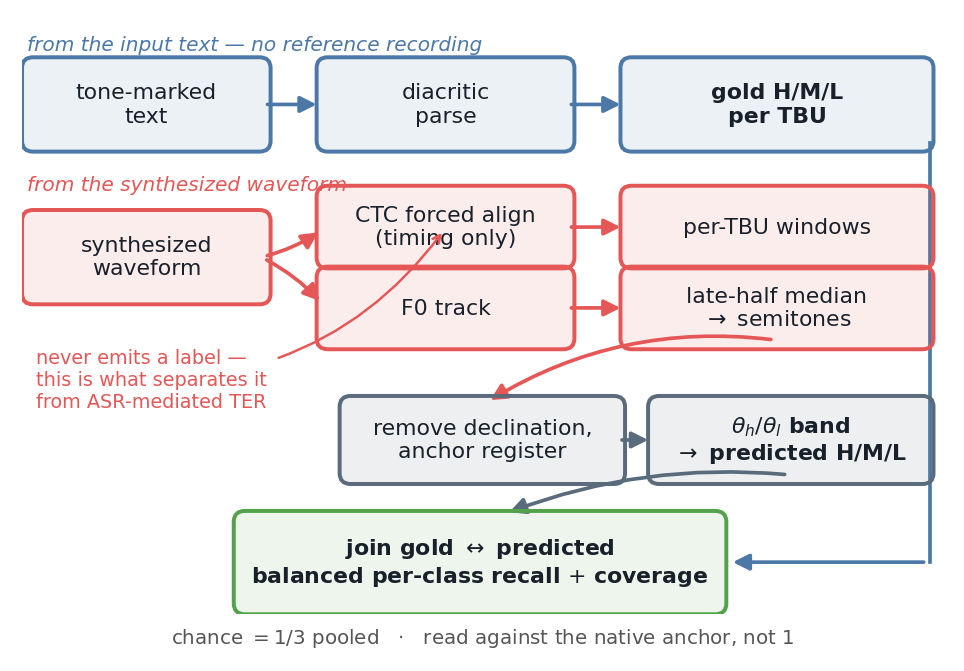}
\caption{\tonemetric{}. Gold comes from the transcript's diacritics; forced alignment supplies one window per
tone-bearing unit and never a label; the declination-removed, register-anchored F0 residual is banded into
H/M/L. The score is a mean of per-class recalls (chance $1/3$ pooled, and at most $1/k$ on a readout holding
only $k$ of the tones), reported with coverage and read against \ANCHOR{}, not $1$.}
\label{fig:dundun}
\end{figure}

\section{Validating \tonemetric{}}
\paragraph{}
Three controls test whether \tonemetric{} tracks tone rather than something correlated with it. Each is built so that a tone-blind score --- one reading intelligibility, quality, or speaker identity --- cannot pass it by accident.

\paragraph{Flattening pitch moves the metric; CER does not.}
PSOLA resynthesis \citep{moulines1990psola} compresses F0 excursions toward a monotone contour by a factor from $1.0$ (untouched) to $0.0$ (fully flat), leaving segmental content, duration and speaker largely intact --- a manipulation aimed precisely at the signal \tonemetric{} claims to read. Across the five settings $\{1.0, 0.75, 0.5, 0.25, 0.0\}$, \tonemetric{} reads $0.543$, $0.549$, $0.518$, $0.372$, $0.333$: it tracks the flattening factor at $r = +0.92$ and lands at $0.333$ --- the three-way chance floor exactly --- once pitch is gone. The ordering is not strictly monotone, the two least-flattened settings being separated by $0.006$ and within run-to-run noise of each other; the fall is unambiguous from $0.5$ downward. CER over the identical conditions does not follow: $0.084$, $0.080$, $0.073$, $0.081$, $0.110$. The result is a \emph{dissociation}, and its direction is easy to overstate: CER does not fall, it stays flat and if anything edges \emph{up} at full flattening --- a small rise we read no signal into. One variable was moved on fixed utterances, and a degradation that destroys lexical tone is total for \tonemetric{} and invisible to CER.

\paragraph{Inverting the answer key.}
The second control changes nothing about the audio or the model. We score 300 native recordings from 37 speakers --- clips no trained model is evaluated on --- against the correct tone key and then against corrupted keys, holding the predicted tone sequences \emph{byte-identical}; only the gold changes. Table~\ref{tab:flip} gives the five conditions. Scrambling the key drives the score to $0.34$, indistinguishable from the $1/3$ chance floor. The High$\leftrightarrow$Low swap is sharper: restricted to the two tones it exchanges, the score reads $0.14$ against a chance level of $0.35$, not $0.50$: the readout restricts the \emph{gold} to High and Low and not the prediction, so a Mid answer there scores as an error and a label-independent predictor is expected at $(1-q_M)/2$, with $q_M{=}0.30$ \tonemetric{}'s own rate of answering Mid. We want to be exact about what this does and does not establish, because the symmetry is not a finding. Over the two exchanged classes the three outcomes partition: $\mathrm{correct} + \mathrm{swapped} + q_M = 1$, so $\mathrm{swapped} = 2\,\mathrm{chance} - \mathrm{correct}$ identically. Substituting the correct-key recalls gives $2(0.349) - 0.558 = 0.140$, which is the measured value to three decimals. The swapped number is therefore \emph{derivable} from the correct one and the Mid rate; it is a consistency check that the scoring path depends on the gold key in the way the definition says, and it would fail loudly if predictions leaked from the answer sheet or if a class were being silently dropped. It is not independent evidence that \tonemetric{} reads pitch. That evidence comes from the previous control, where the audio changes and the key does not. Paired within clip, the three-class drop from the correct to the swapped key is $0.28$ (95\% CI $[0.26, 0.30]$). The correct-key level is itself the native-speech anchor, \ANCHOR{}, and it bounds the scale in practice --- \tonemetric{} does not reach $1.0$ on correctly-toned native speech, so the usable range is far narrower than $[1/3, 1]$ and scores are read against \ANCHOR{}, not against $1$. The oracle is a within-clip contrast on identical predictions and does not depend on that absolute level.

\begin{table}[t]
\centering
\begin{tabular}{lc}
\toprule
Answer key (predictions unchanged) & \tonemetric{} \\
\midrule
Correct, 3-class H/M/L            & $0.596$ \\
High$\leftrightarrow$Low swap, 3-class & $0.32$ \\
Scrambled tones, 3-class           & $0.34$ \\
\emph{chance, 3-class}             & \emph{$0.333$} \\
\midrule
Correct, 2-class H/L               & $0.56$ \\
High$\leftrightarrow$Low swap, 2-class & $0.14$ \\
\emph{chance, 2-class}             & \emph{$0.35$} \\
\bottomrule
\end{tabular}
\caption{Answer-key flip oracle on 300 native recordings (37 speakers). Audio and predicted tone sequences are byte-identical across rows; only the gold key changes. The two-class rows restrict the \emph{gold} to High and Low, not the prediction, so their chance level is $(1-q_M)/2 = 0.35$ at the metric's Mid rate $q_M{=}0.30$, not $1/2$. The swapped readout sits far \emph{below} it --- what a tone-blind score cannot imitate.}
\label{tab:flip}
\end{table}

\paragraph{Human A/B listening --- and what it does not yet establish.}
Three native \yoruba{} listeners judged artifact-matched pairs built by the same PSOLA route from read-speech corpora \citep{gutkin2020yoruba,meyer2022bibletts}: the correctly-toned twin and a tone-flipped twin of one sentence, correct side randomised per trial, rater blind, with five catch trials each. All three cleared the 80\% catch gate ($5/5$, $4/5$, $5/5$). Pooled over 67 scoreable trials, listeners chose the correct twin 60 times --- $89.6\%$, Wilson 95\% CI $[80.0, 94.8]$, exact binomial $p < 10^{-4}$ against $50\%$. The listeners are not equally accurate: two made no errors at all on the non-catch trials and the third made seven. Two raters at ceiling is itself a limitation we return to below, since a rater who is never wrong contributes nothing to an agreement statistic. Independently, with no human input, \tonemetric{} as shipped scored the correct twin above the flipped one in 15 pairs, below in 4, with 5 ties (sign test on the 19 untied pairs, $p = 0.019$). Given a register anchor frozen from the clean twin instead of estimated per utterance --- an oracle setting, and so an upper bound rather than a deployable configuration --- it separates 19 pairs correctly, 3 incorrectly and ties 2 (tie-adjusted win rate $83.3\%$; sign test on the 22 untied pairs, $p = 0.0009$). The gap between the two is itself informative: most of what the shipped metric loses on this task it loses to estimating the speaker's mid-register from a single short utterance, not to reading pitch. Listeners hear the flips, and the metric detects them.

The \emph{per-trial correspondence} between the two, however, is not established at this sample size, and we report that as a negative result rather than a caveat. AUROC of the metric's per-pair margin against human correctness is $0.612$ with 95\% CI $[0.343, 0.879]$: the interval contains $0.50$. The point-biserial correlation is $0.152$, CI $[-0.076, 0.396]$, containing $0$. Adding a third listener moved the point estimate by less than $0.01$ and left the interval straddling chance, so this is not a sample that a few more raters would obviously rescue. We therefore cannot claim that \tonemetric{} finds hard the trials listeners find hard. An earlier single-listener pilot gave a far more optimistic AUROC of $0.96$, but on a superseded stimulus build --- only 7 of its 24 pairs survive into the present set, at different margins --- so the two are not comparable and we report only the present figure. Two further limits compound this: three listeners is a small sample for an agreement statistic, and only one rater individually reached our scoring kit's threshold of 24 scoreable trials (the three contributed 24, 22 and 21), so the estimate rests largely on pooling. The study supports the claim that the flips are audible and that the metric is sensitive to them; whether that sensitivity aligns with a listener's trial by trial is open. Nor is more of the same design obviously the fix: two of our three raters were at ceiling, so their trials carry no variance for an agreement statistic to work with, and a pool of similarly expert listeners would add trials without adding discrimination. Resolving it needs stimuli graded in difficulty rather than uniformly salient --- flips near the threshold of audibility, where listeners actually disagree.

\section{Case Study: What a Few Hours Buys in \yoruba{}}

We fine-tuned the base model \citep{zhu2026omnivoice} on clean, reliably tone-marked \yoruba{} drawn from OpenSLR-86 \citep{gutkin2020yoruba} and BibleTTS \citep{meyer2022bibletts} --- the two public \yoruba{} corpora whose transcripts carry the diacritics \tonemetric{} reads as gold --- at five budgets: $0$ (zero-shot), $1$, $5$, $15$ and $28.5$ hours. One recipe throughout --- ten fixed epochs at $16{,}384$ batch tokens, learning rate $10^{-5}$ --- with one exception, the $28.5$\,h run: the whole clean corpus, $10{,}731$ clips, trained for a fixed $5000$ steps instead (Table~\ref{tab:budget}), which makes it an off-axis point no claim below rests on. Every checkpoint is scored on the same held-out 27-utterance probe, disjoint from the fine-tuning material, along three axes: character error rate (CER) from an MMS recognizer \citep{pratap2023mms}; \tonemetric{}; and speaker similarity (SSIM), the cosine between ECAPA-TDNN speaker embeddings \citep{desplanques2020ecapa} of the generated clip and the prompt speaker's reference, included not as a quality axis but as a control --- a model that improved CER by drifting away from the requested voice would show it here.

\paragraph{Two error bars, which must never share a column.}
A budget sweep has two independent sources of variance, and they answer different questions. Resampling generation from a single trained checkpoint --- five \emph{decode} seeds --- measures how stable that checkpoint's output is. Retraining from scratch with the data held fixed --- three \emph{training} seeds per budget, evaluation seed fixed --- measures what the budget actually buys. Only the second is the error bar a data-efficiency claim needs, and on this sweep it is roughly three to four times the width of the first. Zero-shot is the exception by construction: with no training run there is nothing to vary, so $0$\,h can carry decode variance and nothing else. Table~\ref{tab:budget} therefore labels every interval with the seed type that produced it.

\begin{table}[t]
\centering
\small
\setlength{\tabcolsep}{4pt}
\begin{tabular}{lccl}
\toprule
Budget & \tonemetric{} & CER & Bar from \\
\midrule
$0$\,h (zero-shot) & $0.5674 \pm 0.0197$ & $0.0562$ & 5 decode \\
$1$\,h             & $0.560 \pm 0.025$   & $0.036$  & 3 training \\
$5$\,h             & $0.628 \pm 0.014$   & $0.018$  & 3 training \\
$15$\,h            & $0.676 \pm 0.056$   & $0.021$  & 3 training \\
$28.5$\,h$^{\dagger}$ & $0.633 \pm 0.019$ & $0.0220$ & 5 decode \\
\bottomrule
\end{tabular}
\caption{Clean \yoruba{} budget sweep, 27-utterance probe. The last column is load-bearing: decode bars measure generation stability on one checkpoint, training bars what a budget buys across independent fine-tuning runs; the two must never be differenced. The $0$\,h and $28.5$\,h rows admit only the first. The $5$\,h and $15$\,h training intervals overlap, so their ordering is not resolved. $^{\dagger}$$28.5$\,h also departs from the fixed-epoch recipe ($5000$ steps against $\approx\!2600$) and is not comparable with the bracketed rows.}
\label{tab:budget}
\end{table}

\paragraph{Intelligibility improves sharply and early.}
CER falls from $0.0562$ zero-shot to $0.036$ at $1$\,h and $0.018$ at $5$\,h --- about three times fewer character errors --- and then stops falling, reading $0.021$ at $15$\,h and $0.0220$ at the full $28.5$\,h. Speaker similarity, meanwhile, stays flat: inside $0.81$--$0.84$ at every budget, $0.8148$ zero-shot and $0.8265$ at $28.5$\,h. Voice identity is not what a few hours of clean data changes; the recognizability of the words is.

\paragraph{Tone starts near the top of the axis and saturates it within an hour.}
Before a single hour of \yoruba{} audio, the model reads $0.5674$ --- $0.89$ of the way from the $1/3$ chance level to the \ANCHOR{} anchor, the whole range this metric resolves. The five-decode-seed series then normalises to $0.99$, $1.13$, $1.25$ and $1.14$ at $1$, $5$, $15$ and $28.5$\,h. Tone is therefore not flat, travelling about $0.36$ of that range --- but only $0.11$ of it lay below the anchor, the $1$\,h mean is already inside the anchor's interval, and seven-tenths of the travel happens above it. Those are decode-seed means; the $1$\,h training-seed mean in Table~\ref{tab:budget} sits below the zero-shot mean, but the bar types differ and we read no decrease into it. \tonemetric{} coverage stays above $0.94$ in every run, well clear of the $\sim\!0.7$ level at which a score is unreadable, so the saturation is a measurement, not an abstention artifact.

This is the finding, and it is a dissociation rather than a ranking. The base model's own paper reports $21.4\%$ \yoruba{} CER, its worst African language \citep{zhu2026omnivoice}; on our probe, before it has seen an hour of \yoruba{}, its tone already covers $0.89$ of the chance-to-anchor span. A CER-only evaluation of this sweep would have reported a large, early, real win --- and would have said nothing whatever about whether the tones were right. The two axes answer different questions, and only one of them was being asked.

\paragraph{What the sweep does not show.}
Four limits, none of them cosmetic. First, the $5$\,h and $15$\,h training-seed intervals overlap: the sweep's own three-seed analysis records the curve shape beyond $5$\,h as within training noise. We read the endpoints, not the ordering, and we do not claim that $15$\,h beats $5$\,h. Second, the $28.5$\,h run is one training run and off-recipe, so it cannot be compared against the bracketed rows in either direction; it is reported because it is the full corpus, not because it settles the top of the curve. Third, and most important for the metric itself: from $5$\,h onward the measured tone score sits above \ANCHOR{}. That is not super-native tone. It means the axis saturates near the anchor and stops discriminating there --- a limitation of \tonemetric{}, not a property of the model, and the reason this sweep can show that the tone axis is nearly spent before fine-tuning starts while being unable to rank its own top budgets on tone. Fourth, and the limit that most constrains how these numbers should be read: neither bar type here contains \emph{item} variance. Both hold the same $27$ utterances fixed and resample only the decoder or the training run, so both answer ``how stable is this number on \emph{these} sentences'' and neither answers ``what would it be on another $27$''. That second interval is the larger one. The native anchor's clip-level standard deviation is $0.168$ (300 clips), which puts a $95\%$ item-sampling half-width of about $\pm 0.063$ on any $27$-clip estimate --- roughly a quarter of the whole $[1/3, \ANCHOR{}]$ range the metric resolves, and several times either bar we print. So the ordering of adjacent budgets is not resolved by this design at all, and the range-normalised figures above should be read as locating the sweep within its axis, not as separating one budget from its neighbour. What survives that interval is the gap this section is about: the distance from chance to zero-shot is far larger than anything fine-tuning adds. Finally, this is one tonal language and one base model.

\begin{figure}[t]
\centering
\includegraphics[width=\columnwidth]{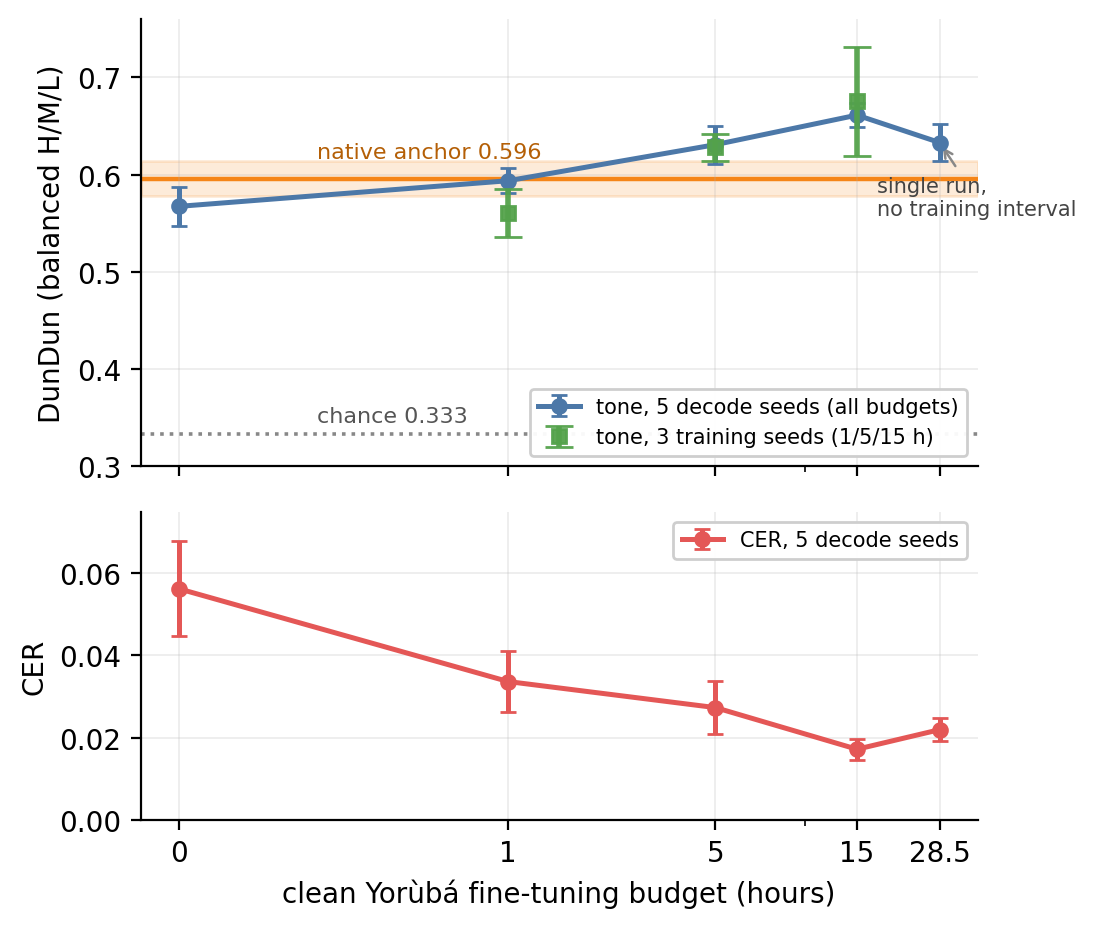}
\caption{The clean \yoruba{} budget curve. \textbf{Top:} \tonemetric{} against the native-speech anchor
(\ANCHOR{}, shaded band the \ANCHORCI{}) and the $1/3$ chance floor. The line carries \emph{decode}-seed
variation --- five generation seeds on one checkpoint, the only protocol covering all five budgets --- and
the squares overlay \emph{training}-seed variation, three independent fine-tuning runs with the data held
fixed, which exists for $1$, $5$ and $15$\,h. The two are different quantities and are deliberately not
merged; $28.5$\,h is a single, off-recipe training run and carries no training interval. \textbf{Bottom:}
character error rate over the same checkpoints. CER falls by roughly threefold, while tone begins at
$0.89$ of the chance-to-anchor span and spends the sweep at or above the anchor, where the axis stops
discriminating.}
\label{fig:budget}
\end{figure}

\section{Generalization: Non-Tonal Swahili}

The same recipe on a language without lexical tone shows what \tonemetric{} is, and is not, for. We fine-tuned the same base model \citep{zhu2026omnivoice} on the Swahili portion of the OpenBible multilingual read-speech collection \citep{openbible_swahili} --- single-narrator studio audio, the non-tonal counterpart of the \yoruba{} material --- under the recipe the \yoruba{} sweep uses --- ten fixed epochs at 16{,}384 batch tokens, seed 4242 --- at budgets of 0, 1, 2.5 and 5 hours.

Character error rate alone carries the result. It falls from $0.0554$ zero-shot to $0.0356$ at 1\,h (345 clips, 91 steps), $0.0422$ at 2.5\,h (849 clips, 225 steps) and $0.0294$ at 5\,h (1714 clips, 453 steps), while speaker similarity stays flat throughout ($0.886$, $0.888$, $0.875$, $0.879$). The 1\,h and 2.5\,h runs invert, which is the expected behaviour of single-seed points a few thousandths apart and a reason to read the endpoints rather than the ordering. A reader holding only the standard metric would draw our conclusion unaided.

That is the argument. \yoruba{} needed one because CER is structurally blind to a contrast that decides word identity there; Swahili does not, because it has no such contrast. What a language needs from an evaluation stack is a property of the language, not a universal upgrade.

The negative claim is narrow. We report no \tonemetric{} score for Swahili and ran no comparison that came out flat: Swahili carries no lexical tone and its orthography marks none, so no gold High/Mid/Low sequence can be read from the input text. The metric is inapplicable here, not uninformative: a flat tone score would have been evidence about the metric, and we have none. This is also one language, one training seed and one run per budget, without error bars. And where \tonemetric{} does apply, its axis runs from $1/3$ only as far as the native-speech anchor \ANCHOR{}: it does not reach $1.0$ on correctly-toned native speech, so the usable range is far narrower than $[1/3,1]$.

\section{Discussion and Limitations}
\paragraph{One tonal language, and a scope condition that is orthographic.}
\tonemetric{} is validated on \yoruba{} alone, and its scope condition is not that a language carries lexical tone but that its orthography marks tone \emph{reliably}: gold High/Mid/Low is read from the input diacritics and from nothing else. Two neighbouring languages fail it. Igbo marks tone only partially in ordinary text, and its downstep --- a lowered High that is neither High nor Mid in register --- has no cell in a three-way key. Hausa orthography does not mark tone at all. \tonemetric{} does not transfer to either as written, and nothing here is evidence that it would; reaching them needs another source of gold --- an expert in the loop \citep{mohanta2026mizo}, or a tone-annotated lexicon --- which is a different method, not a port.

\paragraph{Unmarked Mid biases the key in the flattering direction.}
Inside \yoruba{} the precondition still bites. Mid carries no diacritic, so a gold Mid and a diacritic the transcriber omitted are formally the same event. Under-marking therefore does not add symmetric noise: it turns gold High and Low into gold Mid, flattening the answer key in exactly the direction a flattened synthesizer fails, rewarding a system that never leaves the mid band. Our corpora are curated for tone marking \citep{gutkin2020yoruba,meyer2022bibletts}; over scraped text the same code returns a number that flatters the worst systems most. Diacritized input is a precondition of the measurement, not an assumption that can be relaxed.

\paragraph{The axis saturates before $1.0$.}
Through the identical pipeline, correctly-toned native speech reads \ANCHOR{}, not $1.0$. \tonemetric{} is informative between chance and that anchor and stops ranking systems reliably at or above it --- which is where our own larger budgets land. It can show that most of the axis is spent zero-shot, and cannot resolve the top of its own sweep. Widening that band --- finer thresholds, better register estimation --- is its most valuable improvement.

\paragraph{Three components carry assumptions we have not tested.}
An acoustic model supplies every window by forced alignment \citep{pratap2023mms}, and this is the dominant failure mode: coverage certifies that a window exists, not that it sits on the intended syllable, and a displaced window scores a real F0 value against the wrong gold label, invisibly. That alignment is validated on native read speech and only \emph{assumed} to transfer to synthesized audio. Second, taking the per-utterance median as the register anchor assumes Mid is the modal tone within the clip; an H- or L-heavy carrier drags the anchor into that cluster and shifts every label with it. The frozen-anchor oracle prices this: frozen from a clean twin, it separates $19$ pairs correctly, $3$ wrongly and ties $2$, against $15$ of $24$ as shipped. Third, abstention is not class-independent --- utterance-final Low is often creaky and drops out --- so per-class recall is optimistic wherever it is read without the matching per-class coverage. All three are why the (accuracy, coverage) pair must travel together.

\paragraph{The human evidence is three listeners.}
All three cleared the catch gate ($5/5$, $4/5$, $5/5$), but three is a small sample for an agreement statistic, and only one reached our scoring kit's threshold of $24$ scoreable trials ($24$, $22$, $21$), so the estimate rests on pooling. Per-trial correspondence between metric and listener is unresolved, not established: the AUROC interval contains chance. Two raters made no errors at all on the non-catch trials, which is the deeper problem: a rater at ceiling contributes no variance, so more listeners of this calibre would add trials without adding discrimination. The flips are audible and \tonemetric{} detects them; that it finds hard the trials listeners find hard is not something we can yet claim, and settling it needs stimuli graded in difficulty rather than uniformly salient.

\paragraph{The sweep is narrow.}
It is one model family on one probe. The $5$\,h and $15$\,h training-seed intervals overlap ($0.628 \pm 0.014$ against $0.676 \pm 0.056$), so the curve's shape beyond $5$\,h is within training noise and only its endpoints are read.

\paragraph{What would settle these.}
Three things, in rising cost. More listeners, turning the per-trial question from a negative result into an answer either way. A second tone-marking language, so the scope claim rests on more than one orthography. And an alignment study on synthesized rather than native audio --- the one assumption the codec-independence argument rests on, and the one we have not yet paid for.

\section{Conclusion}
CER cannot see lexical tone, and in \yoruba{} lexical tone is word identity. \tonemetric{} supplies the missing axis, reading gold from the input diacritics and the prediction from the audio's pitch track, so free generations are graded at checkpoint cadence. The axis is narrow --- one tone-marking orthography, a usable range reaching only \ANCHOR{} --- but across a budget sweep it shows a dissociation CER alone would have hidden: hours of clean audio buy intelligibility; tone had spent most of that range before any of them. We release the metric, its frozen calibration, and the validation protocol.

\bibliography{refs}


\end{document}